\RequirePackage{fix-cm}
\documentclass[twocolumn]{svjour3}          
\smartqed  
\usepackage{graphicx}
\usepackage{booktabs}
\usepackage{multirow}
\usepackage{url}
\usepackage{ulem}
\usepackage{amsmath}
\usepackage{subfig}
\usepackage{hyperref}
\begin{document}

\title{Joint Intent Detection and Slot Filling with Wheel-Graph Attention Networks
}


\author{Pengfei Wei \and Bi Zeng \and Wenxiong Liao 
}


\institute{Pengfei Wei \at
              School of Computers, Guangdong University of Technology, Guangzhou, 510006, China \\
              \email{wpf@mail2.gdut.edu.cn}           
           \and
           Bi Zeng \at
             	School of Computers, Guangdong University of Technology, Guangzhou, 510006, China \\
              \email{zb9215@gdut.edu.cn}
           \and
           Wenxiong Liao \at
             	School of Computers, Guangdong University of Technology, Guangzhou, 510006, China \\
              \email{lwx@mail2.gdut.edu.cn}
}

\date{Received: date / Accepted: date}

\maketitle


\begin{abstract}
  Intent detection and slot filling are two fundamental tasks for building a spoken language understanding (SLU) system. Multiple deep learning-based joint models have demonstrated excellent results on the two tasks. In this paper, we propose a new joint model with a wheel-graph attention network (Wheel-GAT) which is able to model interrelated connections directly for intent detection and slot filling. To construct a graph structure for utterances, we create intent nodes, slot nodes, and directed edges. Intent nodes can provide utterance-level semantic information for slot filling, while slot nodes can also provide local keyword information for intent. Experiments show that our model outperforms multiple baselines on two public datasets. Besides, we also demonstrate that using Bidirectional Encoder Representation from Transformer (BERT) model further boosts the performance in the SLU task.
  \keywords{Spoken language understanding \and Graph neural network \and Attention mechanism \and Joint learning}
\end{abstract}

\maketitle

\section{Introduction}
Spoken language understanding (SLU) plays a critical role in the maintenance of goal-oriented dialog systems. The SLU module takes user utterance as input and performs three tasks: domain determination, intent detection, and slot filling \cite{hakkani2016multi}. Among them, the first two tasks are often framed as a classification problem, which infers the domain or intent (from a predefined set of candidates) based on the current user utterance \cite{sarikaya2014application}. For example, the sentence \textit{``play techno on lastfm''} sampled from the SNIPS corpus is shown in Table \ref{tab:1}. It can be seen that each word in the sentence corresponds to one slot label, while a specific intent is assigned for the whole sentence.

\begin{table}
\centering
  \begin{tabular}{|l|c|c|c|c|}
    \hline
    \textbf{Sentence} & play & techno & on & lastfm\\
    \hline
    \textbf{Slots} & O & B-genre & O & B-service\\
    \hline
    \textbf{Intent} & \multicolumn{4}{|c|}{PlayMusic}\\
    \hline
  \end{tabular}
  \caption{An example with intent and slot annotation (BIO format), which indicates the slot of movie name from an utterance with an intent \textit{PlayMusic}.}
  \label{tab:1}
\end{table}

In early research, Intent detection and slot filling are usually carried out separately, which is called traditional pipeline methods. Intent detection is regarded as an utterance classification problem to predict an intent label, which can be modeled using conventional classifiers, including regression, support vector machine (SVM) \cite{haffner2003optimizing} or recurrent neural network (RNN) \cite{lai2015recurrent}. The slot filling task can be formulated as a sequence labeling problem, and the most popular approaches with good performances are conditional random field (CRF) \cite{raymond2007generative} and long short-term memory (LSTM) networks \cite{yao2014spoken}.

Considering this strong correlation between the two tasks, the tendency is to develop a joint model \cite{guo2014joint,liu2016attention,liu2016joint,zhang2016joint}. However, all these models only applied a joint loss function to link the two tasks implicitly. \cite{hakkani2016multi} introduce an RNN-LSTM model where the explicit relationships between the intent and slots are not established. Subsequently, \cite{goo2018slot}, \cite{chen2019self}, and \cite{li2018self} proposed the gate/mask mechanism to explore incorporating the intent information for slot filling. \cite{qin2019stack} adopt the token-level intent detection for the Stack-Propagation framework, which can directly use the intent information as input for slot filling. Recently, some work begins to model the bi-directional interrelated connections for the two tasks. \cite{zhang2019joint} proposed a capsule-based neural network model that accomplishes slot filling and intent detection via a dynamic routing-by-agreement schema. \cite{haihong2019novel} proposed an SF-ID network to establish direct connections for the two tasks to help them promote each other mutually. 

We apply the proposed approach to ATIS and SNIPS datasets from \cite{coucke2018snips} and \cite{goo2018slot}, separately. Our experiments show that our approach outperforms multiple baselines. We further demonstrate that using BERT representations \cite{devlin2019bert} boosts the performance a lot. The contributions of this paper can be summarized as follows: (1) Establishing the interrelated mechanism among intent nodes and slot nodes in an utterance by a graph attention neural network (GAT) structure. (2) We establish a novel wheel graph to incorporate better the semantic knowledge and make our joint model more interpretable. (3) Showing the effectiveness of our model on two benchmark datasets. (4) We examine and analyze the effect of incorporating BERT in SLU tasks.


\section{Related Works}
In this section, we will introduce the related works about SLU and GNN in detail.

\subsection{Spoken Language Understanding}

\textbf{Separate Model} The intent detection is formulated as a text classification problem. The traditional method is to employ n-grams as features with generic entities, such as locations and dates \cite{zhang2016joint}. This type of approach is restricted to the dimensionality of the input space. Another line of popular approaches is to train machine learning models on labeled training data, such as support vector machine (SVM) and Adaboost \cite{haffner2003optimizing,schapire2000boostexter} . Approaches based on deep neural network technology have shown excellent performance, such as Deep belief networks (DBNs) and RNNs \cite{ravuri2015recurrent,deoras2013deep}. Slot filling can be treated as a sequence labeling task. The traditional method based on conditional random fields (CRF) architecture, which has a strong ability on sequence labeling tasks \cite{raymond2007generative}. Another line of popular approaches is CRF-free sequential labeling. \cite{yao2014spoken} introduced LSTM architecture for this task and obtained a marginal improvement over RNN. \cite{shen2018disan} and \cite{tan2018deep} introduce the self-attention mechanism for slot filling.

~\\
\textbf{Implicit Joint Model} Recently, there have been some joint models to overcome the error propagation caused by the pipelined approaches, and all these models only applied share parameters a joint loss function to link the two tasks implicitly. \cite{hakkani2016multi} proposed an RNN-LSTM architecture for joint modeling of intent detection and slot filling. \cite{zhang2016joint} first proposed the joint work using RNNs for learning the correlation between intent and semantic slots of a sentence. \cite{liu2016attention} proposed an attention-based neural network model for joint intent detection and slot filling, which further explores different strategies in incorporating this alignment information into the encoder-decoder framework. All these models outperform the pipeline models by mutual enhancement between two tasks. However, these joint models didn’t model their correlation.

~\\
\textbf{Unidirectional related Joint Model} Recently, some works have explored unidirectional related joint models. These models have exploited the intent information for slot filling. \cite{li2018self} proposed a novel intent-augmented gate mechanism to utilize the semantic correlation between intent and slots fully. \cite{goo2018slot} proposed a slot gate that focuses on learning the relationship between intent and slot attention vectors to obtain better semantic frame results by global optimization. \cite{chen2019bert} utilize a mask gating mechanism to model the relationship between intent detection and slot filling. \cite{qin2019stack} perform the token-level intent detection for the Stack-Propagation framework to better incorporate the intent information.

~\\
\textbf{Interrelated Joint Model} Considering this strong correlation between the two tasks, interrelated joint models have been explored recently.  \cite{wang2018bi} introduce their cross-impact to each other using two correlated bidirectional LSTMs (BLSTM) to perform the intent detection and slot filling tasks jointly. \cite{haihong2019novel} introduce an SF-ID network to establish direct connections for two tasks to help them promote each other mutually. \cite{zhang2019joint} proposed a capsule-based neural network that models hierarchical relationships among word, slot, and intent in an utterance via a dynamic routing-by-agreement schema.

\subsection{Graph Neural Networks}

Applying graph neural networks (GNN) to solve some problems has been a popular approach recently in social network analysis \cite{hamilton2017inductive}, knowledge graphs \cite{hamaguchi2017knowledge}, urban computing, and many other research areas \cite{velivckovic2018graph,huang2019text}. GNN can model non-Euclidean data, while traditional neural networks can only model regular grid data. 

Unlike previously discussed neural network-based methods, our approach explicitly establishes direct connections among intent nodes and slots nodes by GAT \cite{velivckovic2018graph}, which uses weighted neighbor features with feature dependent and structure-free normalization, in the style of attention. Analogous to multiple channels in ConvNet \cite{krizhevsky2012imagenet}, GAT introduces multi-head attention \cite{vaswani2017attention} to enrich the model capacity and to stabilize the learning process. Unlike other models \cite{haihong2019novel,zhang2019joint}, our model does not need to set the number of iterations during training. We have also established a wheel graph structure to learn context-aware information in an utterance better.

\section{Proposed Approaches}

\begin{figure*}[h]
  \centering
  \includegraphics[width=\linewidth]{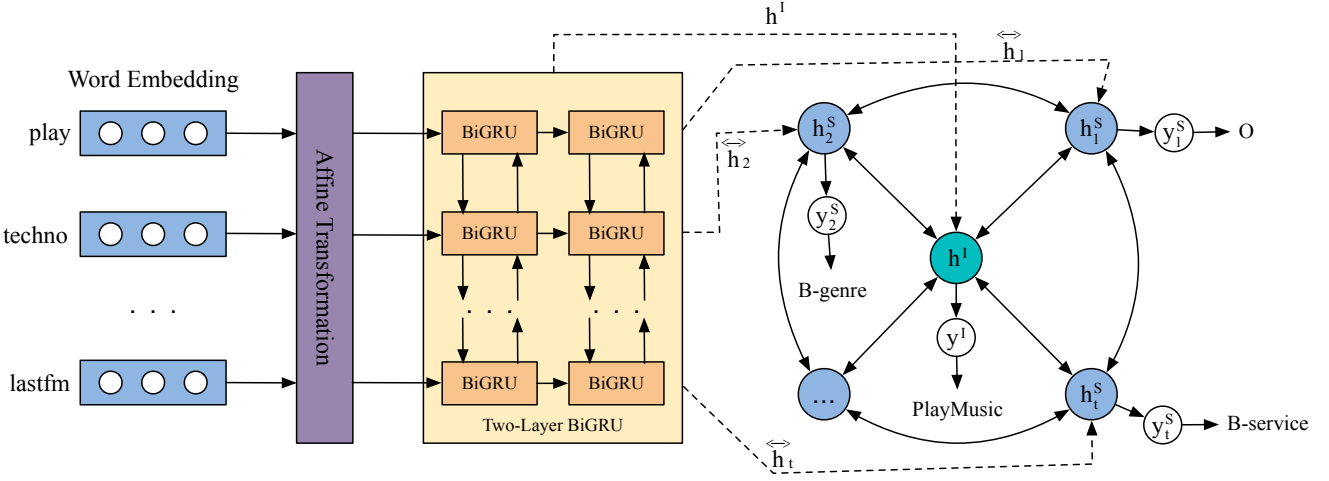}
  \caption{The overall architecture of the proposed model based on Wheel-Graph attention networks.}
  \label{fig:1}       
\end{figure*}

In this section, we will introduce our wheel-graph graph attention model for SLU tasks. The architecture of the model is shown in Figure \ref{fig:1} . First, we show how to uses a text encoder to represent an utterance, which can grasp the shared knowledge between two tasks. Second, we introduce the graph attention network (GAT) user weighted neighbor features with feature dependent and structure-free normalization, in the style of attention. Next, the wheel-graph attention network performs an interrelation connection fusion learning of the intent nodes and slot nodes. Finally, intent detection and slot filling are optimized simultaneously via a joint learning schema.

\subsection{Text Encoder}

\textbf{Word Embedding:} Given a sequence of words, we first covert each word as embedding vector $\mathbf{e}_t$, and the sequence is represented as $[\mathbf{e}_1,…,\mathbf{e}_T]$, where $T$ is the number of words in the sentence. 

~\\
\textbf{Affine Transformation:} We perform an affine transformation on the embedding sequence, which is a data standardization method.

\begin{equation}
  \label{equ:1}
  	\mathbf{x}_t=\mathbf{W}\mathbf{e}_t+\mathbf{b}
\end{equation}
where $\mathbf{W}$ and $\mathbf{b}$ are trainable weights and biases.

~\\
\textbf{Two-Layer BiGRU:} As an extension of conventional feed-forward neural networks, it was difficult to train Recurrent neural networks (RNNs) to capture long-term dependencies because the gradients tend to either vanish or explode. Therefore, some more sophisticated activation functions with gating units were designed.  Two revolutionary methods are long short-term memory (LSTM) \cite{hochreiter1997long} and gated recurrent unit (GRU) \cite{cho2014properties}. Similarly to the LSTM unit, the GRU has gating units that modulate the flow of information inside the unit; however, without having a separate memory cells and has less parameters. Based on this, we use GRU in this work.

\begin{equation}
  \label{equ:2}
  	\mathbf{r}_t=\sigma{(\mathbf{W_r} \mathbf{x}_t+\mathbf{U_r} \mathbf{h}_{t-1})}
\end{equation}
\begin{equation}
  \label{equ:3}
  	\mathbf{z}_t=\sigma{(\mathbf{W_z} \mathbf{x}_t+\mathbf{U_z} \mathbf{h}_{t-1})}
\end{equation}
\begin{equation}
  \label{equ:4}
  	\tilde{\mathbf{h}}_t=tanh(\mathbf{W}\mathbf{x}_t+\mathbf{r_t} \odot (\mathbf{U}\mathbf{h}_{t-1}))
\end{equation}
\begin{equation}
  \label{equ:5}
  	\mathbf{h}_t=(1-\mathbf{z}_t) \odot \mathbf{h}_{t-1}+\mathbf{z}_t \odot \tilde{\mathbf{h}}_t
\end{equation}
where $\mathbf{x}_t$ is the input at time $t$, $\mathbf{r}_t$ and $\mathbf{z}_t$ are reset gate and update gate respectively, $\mathbf{W}$ and $\mathbf{U}$ are weight matrices, $\sigma$ is sigmoid function and $\odot$ is an element-wise multiplication. When the reset gate is off ($\mathbf{r}_t$ close to 0), the reset gate effectively makes the unit act as if it is reading the first symbol of an input sequence, allowing it to forget the previously computed state. For simplification, the above equations are abbreviated with $\mathbf{h}_t=GRU(\mathbf{x}_t,\mathbf{h}_{t-1})$.

To consider both past and future information at the same time. Consequently, we use a two-Layer bidirectional GRU (BiGRU) to learn the utterance representations at each time step. The BiGRU, a modification of the GRU, consists of a forward and a backward GRU. The layer reads the affine transformed output vectors $[\mathbf{x}_1,…,\mathbf{x}_T]$ and generates $T$ hidden states by concatenating the forward and backward hidden states of BiGRU:

\begin{equation}
  \label{equ:6}
  	\overrightarrow{\mathbf{h}}_t=\overrightarrow{GRU}(\mathbf{x}_t,\overrightarrow{\mathbf{h}}_{t-1})
\end{equation}

\begin{equation}
  \label{equ:7}
  	\overleftarrow{\mathbf{h}}_t=\overleftarrow{GRU}(\mathbf{x}_t,\overleftarrow{\mathbf{h}}_{t+1})
\end{equation}

\begin{equation}
  \label{equ:8}
  	\overleftrightarrow{\mathbf{h}}_t=[\overrightarrow{\mathbf{h}}_t,\overleftarrow{\mathbf{h}}_t]
\end{equation}
where $\overrightarrow{\mathbf{h}}_t$ is the hidden state of forward pass in BiGRU, $\overleftarrow{\mathbf{h}}_t$ is the hidden state of backward pass in BiGRU and $\overleftrightarrow{\mathbf{h}}_t$ is the concatenation of the forward and backward hidden states at time $t$.

In summary, to get more fine-grained sequence information, we use a two-layer BiGRU to encode input information. The representation is defined as:

\begin{equation}
  \label{equ:9}
  	\overleftrightarrow{\mathbf{h}}_t=BiGRU(BiGRU(\mathbf{x}_t))
\end{equation}

\subsection{Graph Attention Network}

A graph attention network (GAT) \cite{velivckovic2018graph} is a variant of graph neural network \cite{scarselli2008graph} and is an important element in our proposed method. It propagates the intent or slot information from a one-hop neighborhood. Given a dependency graph with $N$ nodes, where each node is associated with a local vector $\mathbf{x}$, one GAT layer compute node representations by aggregating neighborhood’s hidden states.

GAT exploits the attention mechanism as a substitute for the statically normalized convolution operation. Below are the equations to compute the node embedding $\mathbf{h}_i^{(l+1)}$   of layer $l+1$ from the embeddings of layer $l$.

\begin{equation}
  \label{equ:10}
  	\mathbf{z}_i^{(l)}=\mathbf{W}^{(l)} \mathbf{h}_i^{(l)}
\end{equation}
\begin{equation}
  \label{equ:11}
  	e_{ij}^{(l)}=f(\overrightarrow{\mathbf{a}}^{(l)^T} (\mathbf{z}_i^{(l)}\parallel \mathbf{z}_j^{(l)}))
\end{equation}
\begin{equation}
  \label{equ:12}
  	\alpha_{ij}^{(l)}=\frac{exp(e_{ij}^{(l)})}{\sum_{k\in N(i)}exp(e_{ik}^{(l)})}
\end{equation}
\begin{equation}
  \label{equ:13}
  	\mathbf{h}_i^{(l+1)}=\sigma(\sum_{j\in N(i)}\alpha_{ij}^{(l)}\mathbf{z}_j^{(l)})
\end{equation}
where $\mathbf{W}^{(l)}$ is a linear transformation matrix for input states, $\parallel$ represents vector concatenation, $\overrightarrow{\mathbf{a}}^{(l)}$ is an attention context vector learned during training, and $\cdot^T$ represents transposition. $f(\cdot)$ is a LeakyReLU non-linear function \cite{maas2013rectifier}. $N(i)$ is the neighbor nodes of node $i$. $\sigma$ is the activation function such as $tanh$. For simplification, the above equations are abbreviated with $\mathbf{h}^{(l+1)}=GAT(\mathbf{h}^{(l)})$.

%

\subsection{Wheel-Graph Attention Network}

In the SLU task, there is a strong correlation between intent detection and slot filling. To make full use of the correlation between intent and slot, we constructed a wheel-graph structure. In Figure \ref{fig:1} , this wheel-graph structure contains an intent node and slot nodes.

For the node representation, we use the output of the previous two-layer BiGRU, and the formula is expressed as:

\begin{equation}
  \label{equ:16}
  	\mathbf{h}^I_0=\max_{i=1}^T\overleftrightarrow{\mathbf{h}}_t
\end{equation}
where the max function is an element-wise function, and $T$ is the number of words in the utterance. We use $\mathbf{h}^I_0$ as the representation of the intent node and $\overleftrightarrow{\mathbf{h}}_t$ as the representation of the slot nodes.

For the edge, we created a bidirectional connection between the intent node and the slot nodes. To make better use of the context information of the utterance, we created a bidirectional connection between the slot nodes and connected the head and tail of the utterance to form a loop.

In summary, the feed-forward process of our wheel-graph neural network can be written as:

\begin{equation}
  \label{equ:17}
  	\mathbf{h}_m=[\mathbf{h}^I_0,\overleftrightarrow{\mathbf{h}}_t]
\end{equation}
\begin{equation}
  \label{equ:18}
  	\mathbf{h}_m^{(l+1)}=GRU(GAT(\mathbf{h}_m^{(l)}),\mathbf{h}_m^{(l)})
\end{equation}
\begin{equation}
  \label{equ:19}
  	\mathbf{h}^I,\mathbf{h}_t^S=\mathbf{h}_0^{(l+1)},\mathbf{h}_{1:m}^{(l+1)}
\end{equation}
where $m\in{0,1,…,t}$, $\mathbf{h}^I$ is the hidden state output of the intent, and $\mathbf{h}_t^S$ is the hidden state output of the slots. 

\subsection{Joint Intent Detection and Slot Filling}
The last layer is the output layer. We adopt a joint learning method. The softmax function is applied to representations with a linear transformation to give the probability distribution $\mathbf{y}^I$ over the intent labels and the distribution $\mathbf{y}_t^S$ over the $t-th$ slot labels. Formally,
\begin{equation}
  \label{equ:20}
  	\mathbf{y}^I=softmax(\mathbf{W}^I \mathbf{h}^I+\mathbf{b}^I)
\end{equation}
\begin{equation}
  \label{equ:21}
  	\mathbf{y}_t^S=softmax(\mathbf{W}^S \mathbf{h}_t^S+\mathbf{b}^S)
\end{equation}
\begin{equation}
  \label{equ:22}
  	o^I=argmax(\mathbf{y}^I)
\end{equation}
\begin{equation}
  \label{equ:23}
  	o_t^S=argmax(\mathbf{y}_t^S)
\end{equation}
where $\mathbf{W}^I$ and $\mathbf{W}^S$ are trainable parameters of the model, $\mathbf{b}^I$ and $\mathbf{b}^S$ are bias vectors. $o^I$ and $o_t^S$ are the predicted output labels for intent and slot task respectively.

Then we define loss function for our model. We use $\hat{y}^I$ and $\hat{y}^S$ to denote the ground truth label of intent and slot.

The loss function for intent is a cross-entropy cost function.

\begin{equation}
  \label{equ:24}
  	\mathcal{L}_1=-\sum_{i=1}^{n_I}\hat{y}^{i,I}log(y^{i,I})
\end{equation}

Similarly, the loss function of a slot label sequence is formulated as:

\begin{equation}
  \label{equ:25}
  	\mathcal{L}_2=-\sum_{t=1}^T\sum_{i=1}^{n_S}\hat{y}_t^{i,S}log(y_t^{i,S})
\end{equation}
where $n_I$ is the number of intent label types, $n_S$ is the number of slot label types and $T$ is the number of words in an utterance.

The training objective of the model is minimizing a united loss function:

\begin{equation}
  \label{equ:26}
  	\mathcal{L}_\theta=\alpha\mathcal{L}_1 + (1-\alpha)\mathcal{L}_2
\end{equation}
where $\alpha$ is a weight factor to adjust the attention paid to two tasks.

\section{Experiments}

In this section, we describe our experimental setup and report our experimental results.

\begin{table}[h]
\centering
  \begin{tabular}{l|c|c}
  	\hline
    Datasets & ATIS & SNIPS\\
    \hline
    \# Train & 4,478 & 13,084\\
    \hline
    \# Validation & 500 & 700\\
    \hline
    \# Test & 893 & 700\\
    \hline
    \# Intents & 21 & 7\\
    \hline
    \# Slots & 120 & 72\\
    \hline
    Vocab Size & 722 & 11,241\\
    \hline
    Avg. Length & 11.28 & 9.05\\
    \hline
  \end{tabular}
  \caption{Datasets overview.}
  \label{tab:2}
\end{table}

\subsection{Experimental Setup}
For experiments, we utilize two datasets, including ATIS \cite{hemphill1990atis} and SNIPS \cite{coucke2018snips}, which is collected by Snips personal voice assistant in 2018. They are two public benchmark single-intent datasets, which are widely used as benchmark in SLU research. Compared to the single-domain ATIS dataset, SNIPS is more complicated, mainly due to the intent diversity and large vocabulary. Both datasets used in our paper follows the same format and partition as in \cite{qin2019stack}. The overview of datasets is listed in Table \ref{tab:2}.

To validate the effectiveness of our approach, we compare it to the following baseline approaches. It is noted that the results of some models are directly taken from \cite{qin2019stack}.

\begin{itemize}
  \item \textbf{Joint Seq} applies an RNN-LSTM architecture for slot filling, and the last hidden state of LSTM is used to predict the intent of the utterance \cite{hakkani2016multi}.
  \\
  \item \textbf{Attention BiRNN} adopts an attention-based RNN model for joint intent detection and slot filling. Slot label dependencies are modeled in the forward RNN. A max-pooling over time on the hidden states is used to perform the intent classification \cite{liu2016joint}.
  \\
  \item \textbf{Slot-Gated Full Atten.} utilizes a slot-gated mechanism that focuses on learning the relationship between intent and slot attention vectors. The intent attention context vector is used for the intent classification \cite{goo2018slot}.
  \\
  \item \textbf{Self-Attention Model} first makes use of self-attention to produce a context-aware representation of the embedding. Then a bidirectional recurrent layer takes as input the embeddings and context-aware vectors to produce hidden states. Finally, it exploits the intent-augmented gating mechanism to match the slot label \cite{li2018self}.
  \\
  \item \textbf{Bi-Model} is a new Bi-model based RNN semantic frame parsing network structure which performs the intent detection and slot filling tasks jointly by considering their cross-impact to each other using two correlated bidirectional LSTMs \cite{wang2018bi}.
  \\
  \item \textbf{SF-ID Network} is a novel bi-directional interrelated model for joint intent detection and slot filling. It contains an entirely new iteration mechanism inside the SF-ID network to enhance the bi-directional interrelated connections \cite{haihong2019novel}.
  \\
  \item \textbf{CAPSULE-NLU} introduces a capsule-based neural network model with a dynamic routing-by-agreement schema to accomplish intent detection and slot filling tasks. The output representations of IntentCaps and SlotCaps are used to intent detection and slot filling, respectively \cite{zhang2019joint}.
  \\
  \item \textbf{Stack-Propagation} adopts a Stack-Propagation, which directly uses the intent information as input for slot filling and performs the token-level intent detection to further alleviate the error propagation \cite{qin2019stack}.
\end{itemize}

\subsection{Implementation Details}
In our experiments, the dimensionalities of the word embedding are 1024 for the ATIS dataset and SNIPS dataset. All model weights are initialized with uniform distribution. The number of hidden units of the BiGRU encoder is set as 512. The number of layers of the GAT model is set to 1. Graph node representation is set to 1024. The weight factor $\alpha$ is set to 0.1. We use the Adam optimizer \cite{kingma2014adam} with an initial learning rate of $10^{-3}$, and L2 weight decay is set to $10^{-6}$. The model is trained on all the training data with a mini-batch size of 64. In order to enhance our model to generalize well, the maximum norm for gradient clipping is set to 1.0. We also apply the dropout ratio is 0.2 for reducing overfit. 

We implemented our model using PyTorch\footnote{\url{https://github.com/pytorch/pytorch}} and DGL\footnote{\url{https://github.com/dmlc/dgl}} on a Linux machine with Quadro p5000 GPUs. For all the experiments, we select the model which works the best on the validation set and evaluate it on the test set.

\begin{table*}[h]
\centering
\resizebox{\textwidth}{!}{
\begin{tabular}{l|c|c|c|c|c|c}
\hline
\multirow{2}{*}{Model}                   & \multicolumn{3}{c|}{ATIS Dataset}              & \multicolumn{3}{c}{SNIPS Dataset}             \\ \cline{2-7}
                                         & Slot (F1)     & Intent (Acc)  & Sentence (Acc) & Slot (F1)     & Intent (Acc)  & Sentence (Acc) \\ \hline
Joint Seq \cite{hakkani2016multi}     & 94.3          & 92.6          & 80.7           & 87.3          & 96.9          & 73.2           \\
Attention BiRNN \cite{liu2016joint}    & 94.2          & 91.1          & 78.9           & 87.8          & 96.7          & 74.1           \\
Slot-Gated Full Atten. \cite{goo2018slot} & 94.8          & 93.6          & 82.2           & 88.8          & 97.0          & 75.5           \\
Self-Attentive Model \cite{li2018self}   & 95.1          & 96.8          & 82.2           & 90.0          & 97.5          & 81.0           \\
Bi-Model \cite{wang2018bi}             & 95.5          & 96.4          & 85.7           & 93.5          & 97.2          & 83.8           \\
SF-ID Network \cite{haihong2019novel}           & 95.6          & 96.6          & 86.0           & 90.5          & 97.0          & 78.4           \\
CAPSULE-NLU \cite{zhang2019joint}         & 95.2          & 95.0          & 83.4           & 91.8          & 97.3          & 80.9           \\
Stack-Propagation \cite{qin2019stack}     & 95.9          & 96.9          & 86.5           & 94.2          & 98.0          & 86.9  \\ \hline
Wheel-GAT                                & \textbf{96.0*} & \textbf{97.5*} & \textbf{87.2*}  & \textbf{94.8*} & \textbf{98.4*} & \textbf{87.4*}  \\ \hline
\end{tabular}}
\caption{Comparison results of different methods using Wheel-GAN on ATIS and SNIPS datasets. The numbers with * indicate that the improvement of our model over all baselines is statistically significant with p $<$ 0.05 under t-test.}
\label{tab:3}
\end{table*}

\subsection{Experimental Results}
As with Qin et al \cite{qin2019stack}, we adopt three evaluation metrics in the experiments. For the intent detection task, the accuracy is applied. For the slot filling task, the F1-Score is utilized. Besides, the sentence accuracy is used to indicate the general performance of both tasks, which refers to the proportion of the sentence whose intent and slot are both correctly-predicted in the whole corpus. Table \ref{tab:3} shows the experimental results of the proposed models on ATIS and SNIPS datasets.

We note that the results of unidirectional related joint models are better than implicit joint models like Joint Seq \cite{hakkani2016multi} and Attention BiRNN \cite{liu2016joint}, and the results of interrelated joint models are better than unidirectional related joint models like Slot-Gated Full Atten. \cite{goo2018slot} and Self-Attentive Model \cite{li2018self}. That is likely due to the strong correlation between the two tasks. The intent representations apply slot information to intent detection task while the slot representations use intent information in slot filling task. The bi-directional interrelated model helps the two tasks to promote each other mutually.

We also find that our graph-based Wheel-GAT model performs better than the best prior joint model Stack-Propagation Framework. In ATIS dataset, we achieve 0.6\% improvement on Intent (Acc), 0.1\% improvement on Slot (F1-score) and 0.7\% improvement on Sentence (Acc). In the SNIPS dataset, we achieve 0.4\% improvement on Intent (Acc), 0.6\% improvement on Slot (F1-score), and 0.5\% improvement on Sentence (Acc). This indicates the effectiveness of our Wheel-GAT model. In the previously proposed model, the iteration mechanism used to set the number of iterations is not flexible on training, and the token-level intent detection increases the output load when the utterance is very long. While our model employed graph-based attention network, which uses weighted neighbor features with feature dependent and structure-free normalization, in the style of attention, and directly takes the explicit intent information and slot information further help grasp the relationship between the two tasks and improve the SLU performance.

\begin{table*}[h]
\centering
\resizebox{\textwidth}{!}{
\begin{tabular}{l|c|c|c|c|c|c}
\hline
\multirow{2}{*}{Model}                   & \multicolumn{3}{c|}{ATIS Dataset}              & \multicolumn{3}{c}{SNIPS Dataset}             \\ \cline{2-7}
                                         & Slot (F1)     & Intent (Acc)  & Sentence (Acc) & Slot (F1)     & Intent (Acc)  & Sentence (Acc) \\ \hline
Wheel-GAT                                & 96.0 & 97.5 & 87.2  & 94.8 & 98.4 & 87.4  \\ \hline
Wheel-GAT w/o intent $\rightarrow$ slot              & 95.5          & 97.1          & 86.9           & 93.5          & 98.0          & 85.7           \\
Wheel-GAT w/o slot $\rightarrow$ intent              & 95.4          & 96.8          & 86.6           & 93.9          & 97.9          & 85.8           \\
Wheel-GAT w/o head $\leftrightarrow$ tail                & 95.6          & 97.0          & 86.9           & 94.0          & 97.6          & 85.8           \\
Wheel-GAT w/o GAT                        & 95.0          & 96.2          & 84.3           & 90.8          & 96.7          & 77.6           \\ \hline
\end{tabular}}
\caption{Ablation Study on ATIS and SNIPS datasets. $\rightarrow$ indicates that the intent node points to the edge of the slot node. $\leftarrow$ indicates that the slot node points to the edge of the intent node. $\leftrightarrow$ indicates the edge where the head and tail word nodes are connected in an utterance.}
\label{tab:4}
\end{table*}

\subsection{Ablation Study}
In this section, to further examine the level of benefit that each component of Wheel-GAT brings to the performance, an ablation study is performed on our model. The ablation study is a more general method, which is performed to evaluate whether and how each part of the model contributes to the full model. We ablate four important components and conduct different approaches in this experiment. Note that all the variants are based on joint learning method with joint loss.

\begin{itemize}
  \item Wheel-GAT w/o intent $\rightarrow$ slot, where no directed edge connection is added from the intent node to the slot node. The intent information is not explicitly applied to the slot filling task on the graph layer.
  ~\\
  \item Wheel-GAT w/o slot $\rightarrow$ intent, where no directed edge connection is applied from the slot node to the intent node. The slot information is not explicitly utilized to the intent detection task on the graph layer.
  ~\\
  \item Wheel-GAT w/o head $\leftrightarrow$ tail, where no bidirectional edge connection is used between the intent node and the slot node. We only use joint loss for joint model, rather than explicitly establishing the transmission of information between the two tasks.
  ~\\
  \item Wheel-GAT w/o GAT, where no graph attention mechanism is performed in our model. The message propagation is computed via GCN instead of GAT. GCN introduces the statically normalized convolution operation as a substitute for the attention mechanism.

\end{itemize}

Table \ref{tab:4} shows the joint learning performance of the ablated model on ATIS and SNIPS datasets. We find that all variants of our much model perform well based on our graph structure except Wheel-GAT w/o GAT. As listed in the table, all features contribute to both intent detection and slot filling tasks.

If we remove the intent $\rightarrow$ slot edge from the holistic model, the slot performance drops 0.5\% and 1.3\% respectively on two datasets. Similarly, we remove the slot $\rightarrow$ intent edge from the holistic model, the intent performance down a lot respectively on two datasets. The result can be interpreted that intent information and slot information are stimulative mutually with each other. We can see that the added edge does improve performance a lot to a certain extent, which is consistent with the findings of previous work \cite{goo2018slot,qin2019stack,haihong2019novel} .

If we remove the head $\leftrightarrow$ tail edge from the holistic model, we see 0.4\% drop in terms of F1-score in ATIS and 0.8\% drop in terms of F1-score in SNIPS. We attribute it to the fact that head $\leftrightarrow$ tail structure can better model context-aware information in an utterance.

To verify the effectiveness of the attention mechanism, we remove the GAT and use GCN instead. For GCN, a graph convolution operation produces the normalized sum of the node feature of neighbors. The result shows that the intent performance drops 1.3\% and 1.7\%, the slot performance drops 1.0\% and 4.0\%, and the sentence accuracy drops 2.9\% and 9.8\% respectively on ATIS and SNIPS datasets. We attribute it to the fact that GAT uses weighting neighbor features with feature dependent and structure-free normalization, in the style of attention.

\begin{figure*}[h]
  \centering
  \subfloat[Visualization of the attention weights\\ of slot $\rightarrow$ intent.]{\includegraphics[width=0.5\textwidth]{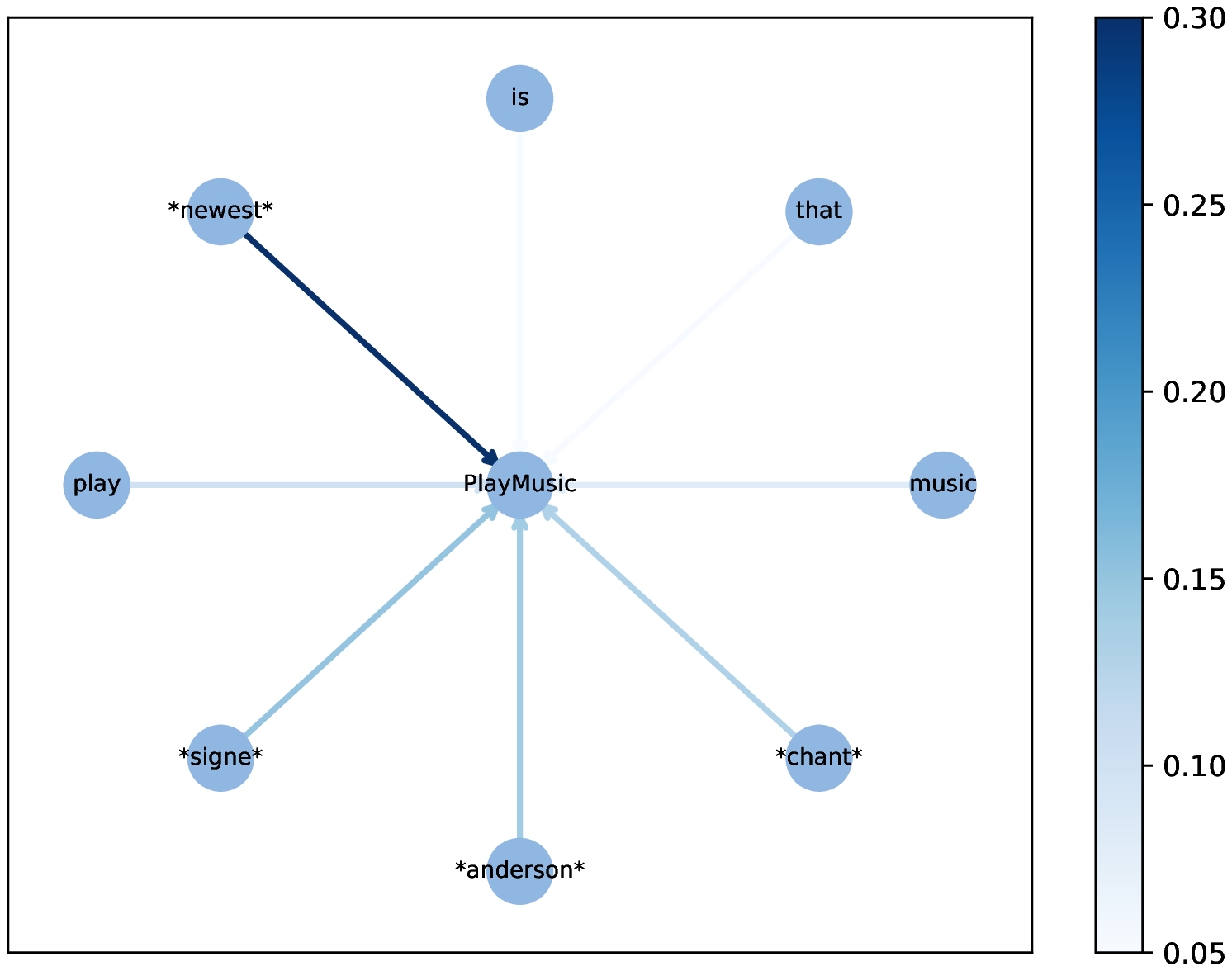} \label{fig:2a}}
	\subfloat[Visualization of the attention weights\\ of each slot node (contains intent $\rightarrow$ slot edges).]{\includegraphics[width=0.5\textwidth]{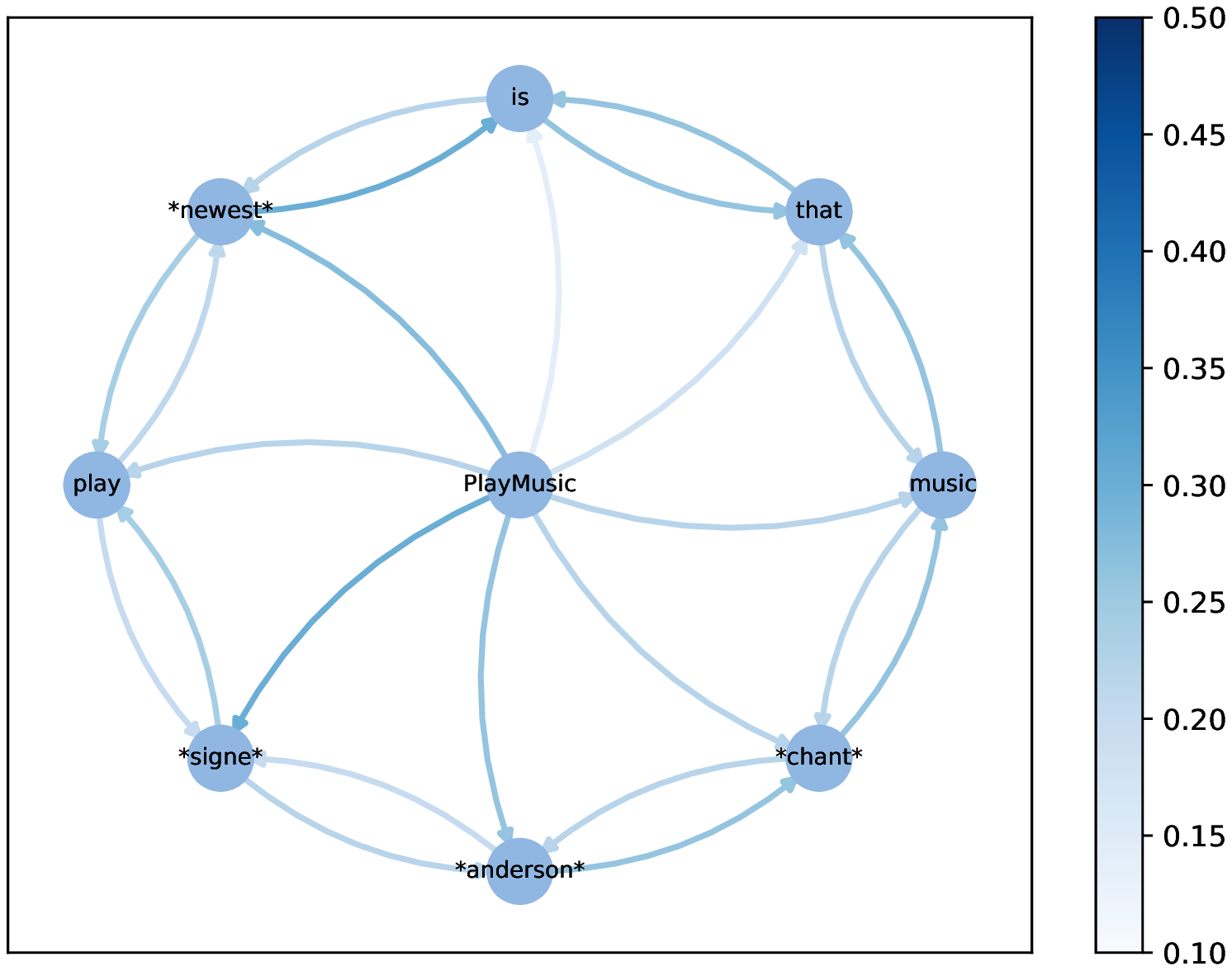} \label{fig:2b}} \\
  \caption{The central node is intent token and slot tokens are surrounded by *. For each edge, the darker the color, it means that this corresponding of the two nodes is more relevant, so that it integrates more information from this source node features.}
	\label{fig:2}
\end{figure*} 

\subsection{Visualization of Wheel-Graph Attention Layer}


In this section, with attempt to better understand what the wheel-graph attention structure has learnt, we visualize the attention weights of slot $\rightarrow$ intent and each slot node, which is shown in Figure \ref{fig:2}. 

Based on the utterance ``\textit{play signe anderson chant music that is newest}'', the intent ``\texttt{PlayMusic}'' and the slot ``\texttt{O B\--artist I\--artist B\--music\_item O O O B\--sort}'', we can clearly see the attention weights successfully focus on the correct slot, which means our wheel-graph attention layer can learn to incorporate the specific slot information on intent node in Figure \ref{fig:2a}. In addition, more specific intent token information is also passed into the slot node in Figure \ref{fig:2b}, which achieves a fine-grained intent information integration for guiding the token-level slot prediction. Therefore, the node information of intent and slots can be transmitted more effectively through attention weights in our proposed wheel-graph attention interaction layer, and promote the performance of the two tasks at the same time.

\begin{table*}[h]
\centering
\resizebox{\textwidth}{!}{
\begin{tabular}{l|c|c|c|c|c|c}
\hline
\multirow{2}{*}{Model}                      & \multicolumn{3}{c|}{ATIS Dataset}         & \multicolumn{3}{c}{SNIPS Dataset}        \\ \cline{2-7} 
                                            & Slot (F1) & Intent (Acc) & Sentence (Acc) & Slot (F1) & Intent (Acc) & Sentence (Acc) \\ \hline
Wheel-GAT                                   & 96.0      & 97.5         & 87.2           & 94.8      & 98.4         & 87.4           \\ \hline
BERT SLU \cite{chen2019bert}                & 96.1      & 97.5         & 88.2           & 97.0      & 98.6         & 92.8           \\
Stack-Propagation + BERT \cite{qin2019stack} & 96.1      & 97.5         & 88.6           & 97.0      & 99.0         & 92.9           \\
Wheel-GAT + BERT                            & 96.5      & 98.0         & 90.2           & 97.4      & 99.3         & 93.6           \\ \hline
\end{tabular}}
\caption{The SLU performance on BERT-based model on ATIS and SNIPS datasets.}
\label{tab:5}
\end{table*}

\subsection{Effect of BERT}

In this section, we also experiment with a pre-trained BERT-based \cite{devlin2019bert} model instead of the Embedding layer, and use the fine-tuning approach to boost SLU task performance and keep other components the same as with our model.

As can be seen from Table \ref{tab:5}, Stack-Propagation + BERT \cite{qin2019stack} joint model achieves a new state-of-the-art performance than another without a BERT-based model, which indicates the effectiveness of a strong pre-trained model in SLU tasks. We attribute this to the fact that pre-trained models can provide rich semantic features, which can help to improve the performance on SLU tasks. Wheel-GAT + BERT outperforms the Stack-Propagation + BERT. That is likely due to we adopt explicit interaction between intent detection and slot filling in two datasets. It demonstrates that our proposed model is effective with BERT.

\section{Conclusion and Future Work}

In this paper, we first applied the graph network to the SLU tasks. And we proposed a new wheel-graph attention network (Wheel-GAT) model, which provides a bidirectional interrelated mechanism for intent detection and slot filling tasks. The intent node and the slot node construct a explicit two-way associated edge. This graph interaction mechanism can provide fine-grained information integration for token-level slot filling to predict the slot label correctly, and it can also provide specific slot information integration for sentence-level intent detection to predict the intent label correctly. The bidirectional interrelated model helps the two tasks promote performance each other mutually. 

We discuss the details of the prototype of the proposed model and introduced some experimental studies that can be used to explore the effectiveness of the proposed method. We first conduct experiments on two datasets ATIS and SNIPS. Experimental results show that our approach outperforms the baselines and can be generalized to different datasets. Then, to investigate the effectiveness of each component of Wheel-GAT in joint intent detection and slot filling, we also report ablation test results in Table \ref{tab:4}. In addition, We visualize and analyze the attention weights of slot $\rightarrow$ intent and each slot node. Besides, we also explore and analyze the effect of incorporating a strong pre-trained BERT model in SLU tasks. Our proposed model achieves the state-of-the-art performance.
 
In future works, our plan can be summarized as follows: (1) We plan to increase the scale of our dataset and explore the efficacy of combining external knowledge with our proposed model. (2) Collecting multi-intent datasets and expanding our proposed model to multi-intent datasets to explore its adaptive capabilities. (3) We plan to introduce reinforcement learning on the basis of our proposed model, and use the reward mechanism of reinforcement learning to improve the performance of the model. (4) Intent detection and slot filling are usually used together, and any task prediction error will have a great impact on subsequent dialog state tracking (DST). How to improve the accuracy of the two tasks while ensuring the stable improvement of the overall evaluation metrics (Sentence accuracy) still needs to be further explored.

\begin{acknowledgements}

This work is supported by the National Natural Science Foundation of China under Grant No.61876043, National Natural Science Foundation of Guangdong Province under Grant No.2018A030313868 and Major Industry-University Research Project of Guangdong Province under Grant No.2016B010108004. The corresponding author of this paper is Bi Zeng.

\end{acknowledgements}

%
%

\bibliographystyle{spmpsci}      
\bibliography{slu}   

%
%

\end{document}